\pdfoutput=1
\documentclass[11pt]{article}

\usepackage[]{ACL2023}

\usepackage{xcolor}
\usepackage{times}
\usepackage{latexsym}
\usepackage{booktabs}
\usepackage{enumitem}

\usepackage[T1]{fontenc}
\usepackage{graphicx}
\usepackage[utf8]{inputenc}

\usepackage{microtype}

\usepackage{inconsolata}

\usepackage{soul}
\usepackage{multirow}
\usepackage{cleveref}

\title{Evaluation of ChatGPT on Biomedical Tasks: A Zero-Shot Comparison with Fine-Tuned Generative Transformers}

 \author{Israt Jahan\textsuperscript{\textdagger, \textdollar}, Md Tahmid Rahman Laskar\textsuperscript{\textdaggerdbl, \textdollar, \textsection}, Chun Peng\textsuperscript{\textdagger}, Jimmy Xiangji Huang\textsuperscript{\textdaggerdbl, \textdollar} \\
 \textsuperscript{\textdagger}Department of Biology, York University\\ \textsuperscript{\textdaggerdbl}School of Information Technology, York University \\
  \textsuperscript{\textdollar}Information Retrieval and Knowledge Management Research Lab, York University \\
  \textsuperscript{\textsection}Dialpad Canada Inc. \\
  Toronto, Ontario, Canada \\
  \texttt{\{israt18,tahmid20,cpeng,jhuang\}@yorku.ca}}

\begin{document}
\maketitle
\begin{abstract}
ChatGPT is a large language model developed by OpenAI\footnote{\url{https://openai.com/blog/chatgpt}}. Despite its impressive performance across various tasks, no prior work has investigated its capability in the biomedical domain yet. To this end, this paper aims to evaluate the zero-shot performance of ChatGPT on various benchmark biomedical tasks, such as relation extraction, document classification, question answering, and summarization. To the best of our knowledge, this is the first work that conducts an extensive evaluation of ChatGPT in the biomedical domain. Interestingly, we find based on our evaluation that in biomedical datasets that have smaller training sets, zero-shot ChatGPT even outperforms the state-of-the-art fine-tuned generative transformer models, such as BioGPT and BioBART. This suggests that ChatGPT's pre-training on large text corpora makes it quite specialized even in the biomedical domain. Our findings demonstrate that ChatGPT has the potential to be a valuable tool for various biomedical tasks that lack large annotated data. 
\end{abstract}
%\vspace{-2mm}
\section{Introduction}
%In recent years, the biomedical domain has witnessed an exponential increase in the volume of published literature, making it increasingly challenging for researchers and practitioners to stay up-to-date with the latest advancements. This information overload has created a pressing need for efficient models that can facilitate the extraction of relevant information from the biomedical literature via relation extraction, question answering, or summarization.  \\
%\vspace{-1mm}
The rapid growth of language models \cite{rogers2021primer,zhou2023comprehensive} in the field of Natural Language Processing (NLP) in recent years has led to significant advancements in various domains, including the biomedical domain \cite{kalyan2022ammu}. %Models capable of processing biomedical literature can support researchers and healthcare professionals in a wide range of tasks, such as, relation extraction, question answering, summarization, etc. 
Although specialized models (e.g., BioBERT \cite{lee2020biobert}, BioBART \cite{yuan2022biobart}, BioGPT \cite{luo2022biogpt}, etc.) have shown promising results in the biomedical domain, they require fine-tuning using domain-specific datasets. This fine-tuning process can be time-consuming due to the requirement of task-specific large annotated datasets. In contrast, zero-shot learning enables models to perform tasks without the need for fine-tuning on task-specific datasets. ChatGPT, a large language model, has demonstrated impressive zero-shot performance across various tasks \cite{laskar2023systematicchatgpt}. However, its performance in the biomedical domain remains to be thoroughly investigated. In this regard, this paper presents a comprehensive evaluation of ChatGPT on four key biomedical tasks: relation extraction, question answering, document classification, and summarization. 

In this paper, our primary objective is to explore the extent to which ChatGPT can perform these tasks without fine-tuning and assess its performance by comparing with state-of-the-art generative fine-tuned models, BioGPT and BioBART. To our best knowledge, this is the first work that evaluated ChatGPT on benchmark biomedical datasets. % Our study contributes to the existing literature by providing insights on the effectiveness of zero-shot learning using large language models  on several biomedical tasks in comparison to the task-specific fine-tuned models. 
Our evaluation of ChatGPT can have a profound impact on the biomedical domain that lacks domain-specific datasets by exploring its zero-shot learning capabilities. To ensure the reproducibility of our evaluation and to help facilitate future research, we will release all the ChatGPT-generated responses along with our evaluation code here: \url{https://github.com/tahmedge/chatgpt-eval-biomed}.  % Our major contributions in this paper are summarized below:

\begin{table*}
\centering
\tiny
\setlength{\tabcolsep}{1.75pt} 
\begin{tabular}{p{1cm}p{1.9cm}p{2.1cm}p{10.3cm}}
\toprule
\textbf{Dataset} & \textbf{Type} & \textbf{Data Split \newline (Train / Valid / Test)} & \textbf{Prompt} \\
\midrule

BC5CDR & Chemical-Disease \newline Relation Extraction & 500 / 500 / 500 & Identify each pair of drugs and the drug-induced side-effects (e.g., diseases)  in the following passage: \newline [PASSAGE]
\\ 
\midrule
KD-DTI & Drug-Target \newline Relation Extraction & 12K / 1K / 1.3K & Identify the drug-target interactions in the following passage (along with the interaction type among the following: 'inhibitor', 'agonist', 'modulator', 'activator', 'blocker', 'inducer', 'antagonist', 'cleavage', 'disruption', 'intercalation', 'inactivator', 'bind', 'binder', 'partial agonist', 'cofactor', 'substrate', 'ligand', 'chelator', 'downregulator', 'other', 'antibody', 'other/unknown'): \newline  [PASSAGE]
\\ 
\midrule
DDI & Drug-Drug \newline Relation Extraction & 664 / 50 / 191 & Identify the pairs of drug-drug interactions in the passage given below based on one of the following interaction types: \newline
(i) mechanism: this type is used to identify drug-drug interactions that are described by their pharmacokinetic mechanism. \newline
(ii) effect: this type is used to identify drug-drug interactions describing an effect. \newline  
(iii) advice: this type is used when a recommendation or advice regarding a drug-drug interaction is given. \newline 
(iv) int: this type is used when a drug-drug interaction appears in the text without providing any additional information. \newline [PASSAGE]
\\ 
\midrule
HoC & Document \newline Classification & 9972 / 4947 / 4947 & The 10 hallmark cancer taxonomy with their definitions are given below: \newline
(i) Sustaining proliferative signaling: Cancer cells can initiate and maintain continuous cell division by producing their own growth factors or by altering the sensitivity of receptors to growth factors. 
(ii) Evading growth suppressors: Cancer cells can bypass the normal cellular mechanisms that limit cell division and growth, such as the inactivation of tumor suppressor genes. 
(iii) Resisting cell death: Cancer cells develop resistance to apoptosis, the programmed cell death process, which allows them to survive and continue dividing.
(iv) Enabling replicative immortality: Cancer cells can extend their ability to divide indefinitely by maintaining the length of telomeres, the protective end caps on chromosomes. 
(v) Inducing angiogenesis: Cancer cells stimulate the growth of new blood vessels from existing vessels, providing the necessary nutrients and oxygen to support their rapid growth.
(vi) Activating invasion and metastasis: Cancer cells can invade surrounding tissues and migrate to distant sites in the body, forming secondary tumors called metastases. 
(vii) Cellular energetics: Cancer cells rewire their metabolism to support rapid cell division and growth, often relying more on glycolysis even in the presence of oxygen (a phenomenon known as the Warburg effect). 
(viii) Avoiding immune destruction: Cancer cells can avoid detection and elimination by the immune system through various mechanisms, such as downregulating cell surface markers or producing immunosuppressive signals. 
(ix) Tumor promoting inflammation: Chronic inflammation can promote the development and progression of cancer by supplying growth factors, survival signals, and other molecules that facilitate cancer cell proliferation and survival. 
(x) Genome instability and mutation: Cancer cells exhibit increased genomic instability, leading to a higher mutation rate, which in turn drives the evolution of more aggressive and drug-resistant cancer cells. \newline
Classify the following sentence in one of the above 10 hallmark cancer taxonomy. If cannot be classified, answer as "empty": \newline [SENTENCE]
%Classification & Classify the sentence given below in one of the 10 categories (i. activating invasion and metastasis, ii. tumor promoting inflammation, iii. inducing angiogenesis, iv. evading growth suppressors, v. resisting cell death, vi. cellular energetics, vii. genomic instability and mutation, viii. sustaining proliferative signaling, ix. avoiding immune destruction, x. enabling replicative immortality') in the Hallmarks of Cancer taxonomy? If cannot be classified, answer as \`empty\'. \newline [SENTENCE]
\\ 
\midrule
PubMedQA & Question Answering & 450 / 50 / 500 & For the question, the reference context, and the answer given below, is it possible to infer the answer for that question from the reference context? Only reply as either Yes or No or Maybe.

Question: [QUESTION]

Reference context:  [REFERENCE CONTEXT]

Answer:  [ANSWER]
\\
\bottomrule
\end{tabular}
%\vspace{-2mm}
\caption{\small{Datasets Details with our Prompts for Relation Extraction, Document Classification, and Question Answering Tasks.}}
\label{tab:input_prompt_biogpt}
\end{table*}

%\vspace{-2mm}
\section{Related Work}
%\vspace{-1mm}
%In this section, we review the prior work in the field of biomedical natural language processing (BioNLP), with a specific focus on ChatGPT and biomedical language models, such as BioBERT, BioElectra, BioGPT, and BioBART. % This survey of related work aims to provide context and motivation for our investigation of ChatGPT's zero-shot capabilities in relation to the fine-tuned BioGPT and BioBART models.

%The recent success of transformer-based language models like BERT and GPT have proven their efficacy in several BioNLP tasks. Alongside these models, sequence-to-sequence models have also emerged as a potent BioNLP tool, exemplified by BioBART.

%Biomedical Natural Language Processing (BioNLP) has seen significant progress in recent years due to 
The effective utilization of transformer-based \cite{vaswani2017attention} NLP models like BERT \cite{devlin2018bert} have also led to significant progress in the biomedical domain \cite{lee2020biobert,clinicalbert,beltagy2019scibert,blurb,blue} in recent years. BERT leverages the encoder of the transformer architecture, while GPT leverages the decoder of the transformer. In addition to these models, sequence-to-sequence models like BART \cite{bart} that leverage both the encoder and the decoder of transformer have also emerged as a powerful approach in various text generation tasks. % By first pre-training these models on a vast amount of text data, task-specific fine-tuning of these models has led to a significant performance gain, outperforming prior works. 

It has been observed that domain-specific pre-training of these models on the biomedical text corpora followed by fine-tuning on task-specific biomedical datasets have helped these models to achieve state-of-the-art performance in a variety of BioNLP tasks \cite{gu2021domain}. However, one major limitation of using such fine-tuned models is that they require task-specific large annotated datasets, which is significantly less available in the BioNLP domain in comparison to the general NLP domain. In this regard, having a strong zero-shot model could potentially alleviate the need for large annotated datasets, as it could enable the model to perform well on tasks that it was not trained on. 

Recently, large autoregressive language models like GPT-3 \cite{gpt3} have demonstrated impressive few-shot learning capability. More recently, a new variant of GPT-3, called the InstructGPT model \cite{ouyang2022training} has been proposed that leverages the reinforcement learning from human feedback (RLHF) mechanism. The resulting InstructGPT models (in other words, GPT-3.5) are much better at following instructions than the original GPT-3 model, resulting in an impressive zero-shot performance across various tasks. ChatGPT, a very recent addition to the GPT-3.5 series, has been trained using dialog-based instructional data alongside its regular training phase. Though ChatGPT has demonstrated strong zero-shot performance across various NLP tasks \cite{laskar2023systematicchatgpt,qin2023chatgpt,bang2023multitaskchatgpt,yang2023exploringchatgpt}, it is yet to be investigated in the biomedical domain. To this end, this paper aims to evaluate ChatGPT in the biomedical domain.% Meanwhile, by analyzing the performance of ChatGPT in BioNLP tasks, this paper will also help accelerate future research with large language models in the biomedical domain. 

\section{Our Methodology}
%\vspace{-2mm}
For a given test sample $X$, we prepare a task instruction $T$ and concatenate the text in the test sample with the task instruction to construct the prompt $P$. Then the prompt $P$ is given as input to ChatGPT (\textit{gpt-3.5-turbo}) to generate the response $R$. In this paper, we evaluate ChatGPT on 4 biomedical tasks across 11 benchmark datasets. Below, we describe these tasks, the datasets we use for evaluation, and the prompt $P$ that we construct for each task depending on the respective dataset.

\begin{table*}
\centering
\tiny
\setlength{\tabcolsep}{1.25pt} 
\begin{tabular}{p{2cm}p{1.5cm}p{2.25cm}p{9.25cm}}
\toprule
\textbf{Dataset} & \textbf{Type} & \textbf{Data Split \newline(Train / Valid / Test)} &  \textbf{Prompt} \\
\midrule
iCliniq & Dialog \newline Summarization & 24851 / 3105 / 3108 & Write a very short and concise one line summary of the following dialogue as an informal question in a healthcare forum:
 \newline [DIALOGUE]
\\ 
\midrule
HealthCare Magic & Dialog \newline  Summarization & 181122 / 22641 / 22642 & Write a very short and concise one line summary of the following dialogue as a question in a healthcare forum:
  \newline [DIALOGUE]
\\ 
\midrule
MeQSum & Question \newline  Summarization & 500 / - / 500 & Rewrite the following question in a short and concise form:   \newline [QUESTION]
\\
\midrule
MEDIQA-QS & Question  \newline Summarization & - / 50 / 100 & Rewrite the following question in a short and concise form:   \newline [QUESTION]
\\

\midrule
MEDIQA-MAS & Answer  \newline Summarization & - / 50 / 80 & For the following question, some relevant answers are given below. Please write down a short concise answer by summarizing the given answers. 

Question: [QUESTION]

Answer 1:  [ANSWER1]

Answer 2:  [ANSWER2]
\\
\midrule
MEDIQA-ANS & Answer \newline  Summarization & - / - / 552 & Write a very short and concise summary of the following article based on the question given below: \newline Question: [QUESTION] \newline Article: [ARTICLE]
\\
\bottomrule

\end{tabular}
%\vspace{-2mm}
\caption{\small{Datasets Details with our Prompts for Abstractive Summarization tasks.}}
\label{tab:input_prompt_summ}
\end{table*}

\paragraph{(i) Relation Extraction:} 
Given a text sequence $S$, the biomedical relation extraction task aims to extract relations between entities mentioned in the text by identifying all possible relation triplets. In this paper, we evaluate drug-target-interaction in the KD-DTI dataset \cite{kddti}, chemical-disease-relation in the BC5CDR dataset \cite{bc5cdr}, and drug-drug-interaction in the DDI dataset \cite{ddi}. Our prompts for these datasets are demonstrated in Table \ref{tab:input_prompt_biogpt}. 
%\\\textbf{Datasets:} We use the following datasets to evaluate ChatGPT on the relation extraction task: \\\textit{(i) BC5CDR: for chemical-disease relation. \newline (ii) KD-DTI: for drug-target interaction extraction. \newline (iii) DDI: for drug-drug interaction extraction}. \newline 
%\\\textbf{Prompt:} We design different prompts for different datasets, as demonstrated in Table \ref{tab:input_prompt_biogpt}.

\paragraph{(ii) Document Classification:} 
%\textbf{Task Description:}  %\textbf{Datasets:}  %\textbf{Prompt:} 
Given a text document $S$, the goal is to classify the type of the document. 
For this task, we use the HoC (the Hallmarks of Cancers corpus) dataset \cite{hoc} that consists of 1580 PubMed abstracts. This dataset was annotated at the sentence level by human experts among ten currently known hallmarks of cancer. 
Our prompt is shown in Table \ref{tab:input_prompt_biogpt}. 

\paragraph{(iii) Question Answering:} 
For the question-answering task, we evaluate the performance of ChatGPT on the PubMedQA dataset \cite{jin2019pubmedqa}. Here, the objective is to determine whether the answer to a given question can be inferred from the reference context. 
We give the question, the reference context, and the answer as input to ChatGPT to determine whether the answer to the given question can be inferred from the given reference context, with ChatGPT being prompted to reply either as \textit{yes}, \textit{no}, or \textit{maybe} (see Table \ref{tab:input_prompt_biogpt} for details).

\paragraph{(iv) Abstractive Summarization:} Given a text sequence $S$, the goal is to generate a concise abstractive summary of $S$. To this end, we evaluate ChatGPT on various biomedical summarization tasks, such as healthcare question summarization (we used \textit{MeQSum} \cite{abacha2019summarization} and \textit{MEDIQA-QS} \cite{abacha2021overview} datasets), medical answer summarization (we used \textit{MEDIQA-ANS} \cite{savery2020questionmediqaans} and \textit{MEDIQA-MAS} \cite{abacha2021overview} datasets), and dialogue summarization (we used the \textit{iCliniq} and \textit{HealthCareMagic} datasets \cite{zeng2020meddialog,mrini2021gradually} for doctor-patient dialogue summarization to generate short queries for healthcare forums describing patient's medical conditions). We show our prompts for this task in Table \ref{tab:input_prompt_summ}.

\begin{table*}
\centering
\tiny
\begin{tabular}{cccccccccccc}
\toprule
%& \multicolumn{11}{c}{\textbf{Dataset}}  \\ 
%\cmidrule(lr){2-12} 
  & \multicolumn{3}{c}{\textbf{BC5CDR}} & \multicolumn{3}{c}{\textbf{KD-DTI}} & \multicolumn{3}{c}{\textbf{DDI}} & \multicolumn{1}{c}{\textbf{HoC}} & \multicolumn{1}{c}{\textbf{PubmedQA}}  \\ 
\cmidrule(lr){2-4}\cmidrule(lr){5-7}\cmidrule(lr){8-10}\cmidrule(lr){11-11}\cmidrule(lr){12-12}
 \textbf{Model} & Precision & Recall & F1 & Precision & Recall & F1 & Precision & Recall & F1 & F1 & Accuracy \\ 
\midrule
\textbf{ChatGPT} & 36.20 & \textbf{73.10} & \textbf{48.42} & 19.19 & \textbf{66.02} & 29.74 & \textbf{47.11} & \textbf{45.77} & \textbf{46.43} & 59.14 & 51.60 \\ 
\textbf{BioGPT} & \textbf{49.52} & 43.25 & 46.17 & \textbf{40.00} & 39.72 & \textbf{38.42} & 41.70 & 44.75 & 40.76 & \textbf{85.12} & \textbf{78.20} \\ 
\bottomrule
\end{tabular}
%\vspace{-2mm}
\caption{\small{Performance on Relation Extraction, Document Classification, and Question Answering datasets.}}
\label{tab:performance_biogpt}
\end{table*}

\begin{table*}
\centering
\tiny
\setlength{\tabcolsep}{3pt} % adjust the column separation

\begin{tabular}{cccccccccccccccccccccccccc}
\toprule
%& \multicolumn{20}{c}{\textbf{Dataset}}  \\ 
%\cmidrule(lr){2-25} 
  & \multicolumn{4}{c}{\textbf{iCliniq}} & \multicolumn{4}{c}{\textbf{HealthCareMagic}}  & \multicolumn{4}{c}{\textbf{MeQSum}}  & \multicolumn{4}{c}{\textbf{MEDIQA-QS}} & \multicolumn{4}{c}{\textbf{MEDIQA-MAS}} & \multicolumn{4}{c}{\textbf{MEDIQA-ANS}}  \\ 
\cmidrule(lr){2-5}\cmidrule(lr){6-9}\cmidrule(lr){10-13}
\cmidrule(lr){14-17}\cmidrule(lr){18-21} \cmidrule(lr){22-25}
{\textbf{Model}} & R-1 & R-2 & R-L & B-S & R-1 & R-2 & R-L & B-S & R-1 & R-2 & R-L & B-S & R-1 & R-2 & R-L & B-S & R-1 & R-2 & R-L & B-S & R-1 & R-2 & R-L & B-S \\ 
\midrule
\textbf{ChatGPT} 
& 30.5 & 12.8 & 25.4 & 89.3
& 28.1 & 9.8 & 24.0  & 88.9
& 30.0 & 12.3 & 26.2  & 89.0
& 30.6 & 11.6 & 26.7  & 89.0
& \textbf{38.9} & \textbf{14.6} & \textbf{22.1} & \textbf{87.9}
& \textbf{28.7} & \textbf{10.4} & \textbf{24.4}  & \textbf{89.0}
 \\ 
\textbf{BioBART-Base} 
& \textbf{61.1} & \textbf{48.5} & \textbf{59.4}  & \textbf{94.1}
& \textbf{46.7} & 26.0 & 44.1  & 91.8
& 53.8 & 36.5 & 51.3  & 92.9
& 30.1 & 11.3 & 27.4  & 89.8
& {32.9} & {11.3} & {29.3}  & 86.1
& 19.0 & 7.5 & 16.8  & 85.0
 \\ 
\textbf{BioBART-Large} 
& 60.3 & 48.0 & 58.7  & 94.0
& 46.5 & \textbf{26.1} & \textbf{44.2}  & \textbf{91.9}
& \textbf{55.6} & \textbf{38.1} & \textbf{53.2}  & \textbf{93.3}
& \textbf{32.0} & \textbf{12.4} & \textbf{29.7}  & \textbf{90.3}
& 30.6 & 10.4 & 27.0  & 86.1
& 21.6 & 9.3 & 19.2  & 85.7
 \\ 
\bottomrule
\end{tabular}
%\vspace{-2mm}
\caption{\small{Performance on Abstractive Summarization datasets. Here, `R-1', `R-2', `R-L' and `B-S' denote `ROUGE-1', `ROUGE-2', `ROUGE-L', and `BERTScore', respectively. For BERTScore \cite{zhang2019bertscore}, we use the RoBERTa-Large model \cite{liu2019roberta}. The results for BioBART models are taken from \citet{yuan2022biobart}.}}
\label{tab:performance_biobart}
\end{table*}

%\vspace{-1mm}
\section{Experiments}
%\vspace{-1mm}
%In this section, we present our experimental findings. 
Since ChatGPT is a generative model, we consider two state-of-the-art generative transformers as our baselines. Below, we first present these baselines, followed by presenting the results. 
%\vspace{-2mm}
\subsection{Fine-tuned Baselines}
%\vspace{-1mm}
\paragraph{BioGPT:} The backbone of BioGPT \cite{luo2022biogpt} is GPT-2 \cite{gpt2}, which is a decoder of the transformer. The BioGPT model was trained over PubMed titles and abstracts via leveraging the standard language modeling task. We compare zero-shot ChatGPT with BioGPT models fine-tuned on relation extraction, document classification, and question-answering tasks. 

\paragraph{BioBART:} BioBART is a sequence-to-sequence model that was pre-trained over PubMed abstracts \cite{yuan2022biobart}. The pre-training process involves reconstructing corrupted input sequences. We compare the zero-shot ChatGPT with BioBART fine-tuned on abstractive summarization datasets. 

%\vspace{-2mm}
\subsection{Results \& Discussion}
%\vspace{-1mm}
We first compare the performance of ChatGPT with BioGPT on relation extraction, document classification, and the question-answering task (see Table \ref{tab:performance_biogpt}). Then we compare its performance with BioBART on summarization datasets (see Table \ref{tab:performance_biobart}). More evaluation details are given in Appendix \ref{sec:appendix_chatgpt_evaluation}. 

\paragraph{Relation Extraction Evaluation:} We observe that in the BC5CDR and KD-DTI datasets for relation extraction, ChatGPT led to higher recall scores but much lower precision scores compared to the fine-tuned BioGPT model. This is because ChatGPT tends to generate long and descriptive responses, leading to many inaccurate relation extractions. Though in terms of F1, it outperforms fine-tuned BioGPT in the BC5CDR dataset, it fails to outperform in the KD-DTI dataset. More importantly, it outperforms BioGPT in the DDI dataset in all metrics: Precision, Recall, and F1. 

While analyzing the results in different datasets, we observe that in both BC5CDR and DDI datasets where ChatGPT outperforms BioGPT, the training set is small, only 500 and 664 instances, respectively. On the other hand, in the KD-DTI dataset where ChatGPT fails to outperform BioGPT, the training set contains 12000 instances. This gives us a strong indication that even in the biomedical domain, zero-shot ChatGPT can outperform fine-tuned biomedical models in smaller-sized datasets.

We also observe that more descriptive prompts may help ChatGPT to obtain better Precision scores. Contrary to the KD-DTI dataset, we describe the definition of each interaction type in the DDI dataset (see Table \ref{tab:input_prompt_biogpt}) where ChatGPT performs the best. To further investigate the effect of prompts in relation extraction, we evaluate the performance in BC5CDR with a new prompt: \newline
{\textit{\textbf{i.} Identify the chemical-disease interactions in the passage given below: [PASSAGE]}}. \newline We observe that the Precision, Recall, and F1 scores are decreased by 16.07\%, 10.3\%, and 14.29\%, respectively, with this prompt variation. 

\paragraph{Document Classification Evaluation:} 
We observe that in the HoC dataset, the zero-shot ChatGPT achieves an F1 score of 59.14, in comparison to its counterpart fine-tuned BioGPT which achieves an F1 score of 85.12. We also investigate the effect of prompt tuning by evaluating with two new prompts that are less descriptive (see Appendix \ref{sec:appendix_prompt_variation} for more details): \newline
 \textit{\textbf{i.} Prompting without explicitly mentioning the name of 10 HoC classes, drops F1 to 38.20.} \newline
 \textit{\textbf{ii.} Prompting with the name of each HoC class is given without providing the definition of each class, drops the F1 score to 46.93.}

\paragraph{Question Answering Evaluation:} 
We observe that in the PubMedQA dataset, the zero-shot ChatGPT achieves much lower accuracy than BioGPT (51.60 by ChatGPT in comparison to 78.20 by BioGPT). However, the BioGPT model was fine-tuned on about 270K QA-pairs in various versions of the PubMedQA dataset for this task. While ChatGPT achieves more than 50\% accuracy even without any few-shot examples in the prompt. 

\paragraph{Summarization Evaluation:} 
We observe that in terms of all ROUGE scores \cite{rouge}, ChatGPT performs much worse than BioBART in datasets that have dedicated training sets, such as iCliniq, HealthCareMagic, and MeQSum. Meanwhile, it performs on par with BioBART in the MEDIQA-QS dataset. More importantly, it outperforms BioBART in both MEDIQA-ANS and MEDIQA-MAS datasets. Note that MEDIQA-ANS, MEDIQA-MAS, and MEDIQA-QS datasets do not have any dedicated training data and ChatGPT achieves comparable or even better performance in these datasets compared to the BioBART model fine-tuned on other related datasets \cite{yuan2022biobart}. This further confirms that zero-shot ChatGPT is more useful than domain-specific fine-tuned models in biomedical datasets that lack large training data.

%While on dialogue summarization datasets that have large training sets , ChatGPT performs much worse than baselines.

% Although ChatGPT performs quite badly in terms of the ROUGE scores in many datasets, it achieves much better performance in all datasets in terms of BERTScore \cite{zhang2019bertscore}. This gives an implication that zero-shot ChatGPT might generate responses by missing vocabularies that appear more frequently in both training and test sets of a dataset, but it still generates a summary that is contextually relevant to the reference summary.
%\vspace{-2mm}
\section{Conclusions and Future Work}
%\vspace{-2mm}
In this paper, we evaluate ChatGPT on 4 benchmark biomedical tasks to observe that in datasets that have large training data, ChatGPT performs quite poorly in comparison to the fine-tuned models (BioGPT and BioBART), whereas it outperforms fine-tuned models on datasets where the training data size is small. These findings suggest that ChatGPT can be useful in low-resource biomedical tasks. We also observe that ChatGPT is sensitive to prompts, as variations in prompts led to a noticeable difference in results. 

Though in this paper, we mostly evaluate ChatGPT on tasks that require it to generate responses by only analyzing the input text, in the future, we will investigate the performance of ChatGPT on more challenging tasks, such as named entity recognition and entity linking \cite{yadav2018surveyner,yan2021unifiedner,yuan2022generativeel,laskar-etal-2022-improving,laskar2022auto,laskar-etal-2022-blink}, as well as problems in information retrieval \cite{huang2005york,huang2009bayesian, yin2010survival, laskar-LREC,laskar-etal-2022-domain}. We will also explore the ethical implications (bias or privacy concerns) of using ChatGPT in the biomedical domain.

\section*{Limitations}
Since the training datasets of ChatGPT are unknown, some data used for evaluation may or may not exist during the training phase of ChatGPT  Also, a new version called the GPT-4 model has been released that may ensure higher accuracy. Nonetheless,  GPT-4 is very costly to use, around 60x more expensive than ChatGPT. Meanwhile, even using the paid ChatGPT Plus\footnote{\url{https://openai.com/blog/chatgpt-plus}, last accessed June 4th, 2023} subscription, it is available for just a limited use (allows evaluation of only 25 samples in 3 hours). 
Another limitation of this research is that the results mentioned in this paper for ChatGPT may not be reproducible, as ChatGPT may generate different responses for the same input prompt. Although the experimental results may change over time, this work will still give a concrete direction for future research using ChatGPT like large language models in the biomedical domain.

%while the model is also continuously updated by OpenAI (note that we used the \textit{gpt-3.5-turbo} model of ChatGPT)

\section*{Ethics Statement}
 The paper evaluates ChatGPT on 4 benchmark biomedical tasks that require ChatGPT to generate a response based on the information provided in the input text. Thus, no data or prompt was provided as input that could lead to ChatGPT generating any responses that pose any ethical or privacy concerns.  This evaluation is only done in some academic datasets that already have gold labels available and so it does not create any concerns like humans relying on ChatGPT responses for sensitive issues, such as disease diagnosis. Since this paper only evaluates the performance of ChatGPT and investigates its effectiveness and limitations, conducting this evaluation does not lead to any unwanted biases. Only the publicly available academic datasets are used that did not require any licensing. Thus, no personally identifiable information has been used. 
 
\section*{Acknowledgements} 
We would like to thank all the anonymous reviewers for their detailed review comments. This work is done at York University and supported by the Natural Sciences and Engineering Research Council (NSERC) of Canada and the York Research Chairs (YRC) program.   % Jimmy Huang (\emph{jhuang@yorku.ca}) is the contact author of this paper. 

\bibliography{anthology,custom}
\bibliographystyle{acl_natbib}

\appendix

\section{Appendix}
\label{sec:appendix}

\subsection{Evaluating ChatGPT on Different Tasks}
\label{sec:appendix_chatgpt_evaluation}
Since ChatGPT generated responses can be lengthy, and may sometimes contain unnecessary information while not in a specific format, especially in tasks that may have multiple answers (e.g., Relation Extraction), it could be quite difficult to automatically evaluate its performance in such tasks by comparing with the gold labels by using just an evaluation script. Thus, for some datasets and tasks, we manually evaluate the ChatGPT generated responses by ourselves and compare them with the gold labels. Below we describe our evaluation approach for different tasks:

\begin{itemize}
    \item \textbf{Relation Extraction:} The authors manually evaluated the ChatGPT generated responses in this task by comparing them with the gold labels. To ensure the reproducibility of our evaluation, we will release the ChatGPT generated responses. 
     \item \textbf{Document Classification:} We created an evaluation script and identifies if the gold label (one of the 10 HoC classes) is present in the ChatGPT generated response. For fair evaluation, we lowercase each character in both the gold label and the ChatGPT generated response. Our evaluation script will be made publicly available to ensure reproducibility of our findings. 
     \item \textbf{Question Answering:} Similar to Document Classification, we also evaluated using an evaluation script that compares the gold label and the ChatGPT generated response (here, we also convert each character to lowercase). The evaluation script will also be made public.
     \item \textbf{Abstractive Summarization}: We used the HuggingFace's Evaluate\footnote{\url{https://huggingface.co/docs/evaluate/index}} library \cite{wolf2020transformers} to calculate the ROUGE scores and the BERTScore for the Abstractive Summarization task evaluation. 
\end{itemize}

\subsection{Effects of Prompt Variation}
\label{sec:appendix_prompt_variation}
We investigate the effects of prompt tuning in the HoC dataset by evaluating the performance of ChatGPT based on the following prompt variations: 

\begin{itemize}
\item Prompting with explicitly defining the 10 HoC classes achieves an F1 score of 59.14 (see Row 1 in Table \ref{tab:prompt_variation_chatgpt}).
    \item Prompting without explicitly mentioning the name of 10 HoC classes, drops F1 to 38.20 (see Row 2 in Table \ref{tab:prompt_variation_chatgpt}).
    \item Prompting with the name of each HoC class is given without providing the definition of each class, drops the F1 score to 46.93 (see Row 3 in Table \ref{tab:prompt_variation_chatgpt}).
\end{itemize}

Our findings demonstrate that more descriptive prompts yield better results. 

\subsection{Sample ChatGPT Generated Responses}
\label{sec:appendix_generated_responses}
Some sample prompts with the ChatGPT generated responses for Relation Extraction, Document Classification, and Question Answering tasks are given in Table \ref{tab:sample_prompt_response_biogpt} and for the Abstractive Summarization task are given in Table \ref{tab:sample_prompt_response_summ}.

\begin{table*}
\centering
\small
\resizebox{.95\textwidth}{!}{
\begin{tabular}{p{1.5cm}p{10cm}p{1.5cm}}
\toprule
\textbf{\#} & \textbf{Prompt} & \textbf{F1} \\
\midrule
1. & The 10 hallmark cancer taxonomy with their definitions are given below: \newline
(i) Sustaining proliferative signaling: Cancer cells can initiate and maintain continuous cell division by producing their own growth factors or by altering the sensitivity of receptors to growth factors. \newline 
(ii) Evading growth suppressors: Cancer cells can bypass the normal cellular mechanisms that limit cell division and growth, such as the inactivation of tumor suppressor genes.  \newline
(iii) Resisting cell death: Cancer cells develop resistance to apoptosis, the programmed cell death process, which allows them to survive and continue dividing. \newline
(iv) Enabling replicative immortality: Cancer cells can extend their ability to divide indefinitely by maintaining the length of telomeres, the protective end caps on chromosomes.  \newline
(v) Inducing angiogenesis: Cancer cells stimulate the growth of new blood vessels from existing vessels, providing the necessary nutrients and oxygen to support their rapid growth. \newline
(vi) Activating invasion and metastasis: Cancer cells can invade surrounding tissues and migrate to distant sites in the body, forming secondary tumors called metastases. \newline
(vii) Cellular energetics: Cancer cells rewire their metabolism to support rapid cell division and growth, often relying more on glycolysis even in the presence of oxygen (a phenomenon known as the Warburg effect). \newline
(viii) Avoiding immune destruction: Cancer cells can avoid detection and elimination by the immune system through various mechanisms, such as downregulating cell surface markers or producing immunosuppressive signals. \newline
(ix) Tumor promoting inflammation: Chronic inflammation can promote the development and progression of cancer by supplying growth factors, survival signals, and other molecules that facilitate cancer cell proliferation and survival. \newline
(x) Genome instability and mutation: Cancer cells exhibit increased genomic instability, leading to a higher mutation rate, which in turn drives the evolution of more aggressive and drug-resistant cancer cells. \newline

Classify the following sentence in one of the above 10 hallmark cancer taxonomy. If cannot be classified, answer as "empty": \newline 

[SENTENCE]
& 59.14 \\

\midrule

2. & Is it possible to classify the following sentence in one of the 10 categories in the Hallmarks of Cancer taxonomy? If possible, write down the class. \newline

[SENTENCE] 
& 38.20 \\
\midrule
3. & Classify the sentence given below in one of the 10 categories (i. activating invasion and metastasis, ii. tumor promoting inflammation, iii. inducing angiogenesis, iv. evading growth suppressors, v. resisting cell death,vi. cellular energetics, vii. genomic instability and mutation, viii. sustaining proliferative signaling, ix. avoiding immune destruction, x. enabling replicative immortality) in the Hallmarks of Cancer taxonomy? If cannot be classified, answer as ``empty''. \newline

[SENTENCE] 
& 46.93 \\ 

\bottomrule
\end{tabular}}
%\vspace{-2mm}
\caption{\small{Effects of Prompt Variation in the HoC dataset for the Document Classification Task.}}
\label{tab:prompt_variation_chatgpt}
\end{table*}

\begin{table*}
\centering
\tiny
\resizebox{.98\textwidth}{!}{
\setlength{\tabcolsep}{2.5pt} 
\begin{tabular}{p{1cm}p{2cm}p{7cm}p{2.5cm}p{2.5cm}}
\toprule
\textbf{Dataset} & \textbf{Type} & \textbf{Prompt} & \textbf{ChatGPT Response} & \textbf{Gold Label} \\
\midrule

BC5CDR & Chemical-Disease \newline Relation Extraction & Identify each pair of drugs and the drug-induced side-effects (e.g., diseases)  in the following passage: \newline Famotidine-associated delirium. A series of six cases.        Famotidine is a histamine H2-receptor antagonist used in inpatient settings for prevention of stress ulcers and is showing increasing popularity because of its low cost. Although all of the currently available H2-receptor antagonists have shown the propensity to cause delirium, only two previously reported cases have been associated with famotidine. The authors report on six cases of famotidine-associated delirium in hospitalized patients who cleared completely upon removal of famotidine. The pharmacokinetics of famotidine are reviewed, with no change in its metabolism in the elderly population seen. The implications of using famotidine in elderly persons are discussed. & 

Drug: Famotidine \newline
Drug-induced disease: Delirium \newline 
Side effect: Increased risk of delirium in elderly patients & chemical: famotidine
\newline disease: delirium

  \\
\midrule
KD-DTI & Drug-Target \newline Relation Extraction & Identify the drug-target interactions in the following passage (along with the interaction type among the following: 'inhibitor', 'agonist', 'modulator', 'activator', 'blocker', 'inducer', 'antagonist', 'cleavage', 'disruption', 'intercalation', 'inactivator', 'bind', 'binder', 'partial agonist', 'cofactor', 'substrate', 'ligand', 'chelator', 'downregulator', 'other', 'antibody', 'other/unknown'): 
Analogue inhibitors by modifying oseltamivir based on the crystal neuraminidase structure for treating drug-resistant H5N1 virus.        The worldwide spread of H5N1 avian influenza and the increasing reports about its resistance to the existing drugs have made a priority for the development of the new anti-influenza molecules. The crystal structure of H5N1 avian influenza neuraminidase reported recently by Russell et al. [R.J. Russell, L.F. Haire, D.J. Stevens, P.J. Collins, Y. P. Lin, G.M. Blackburn, A.J. Hay, S.J. Gamblin, J.J. Skehel, The structure of H5N1 avian influenza neuraminidase suggests new opportunities for drug design, Nature 443 (2006) 45-49] have provided new opportunities for drug design in this regard. It is revealed through the structure that the active sites of the group-1 neuraminidases, which contain the N1 subtype, have a very different three-dimensional structure from those of group-2 neuraminidases. The key difference is in the 150-loop cavity adjacent to the conserved active site in neuraminidase. Based on these findings and by modifying oseltamivir, six analog inhibitors were proposed as candidates for developing inhibitors against H5N1 virus, particularly against the oseltamivir-resistant H5N1 virus strain.  &  

Drug-target interactions: \newline
- Oseltamivir and neuraminidase \newline
- Analog inhibitors and neuraminidase \newline

Interaction type: Inhibitor \newline & drug: Oseltamivir \newline
target: Neuraminidase \newline 
interaction: inhibitor

\\ 
\midrule
DDI & Drug-Drug \newline Relation Extraction & Identify the pairs of drug-drug interactions in the passage given below based on one of the following interaction types: \newline
(i) mechanism: this type is used to identify drug-drug interactions that are described by their pharmacokinetic mechanism. \newline
(ii) effect: this type is used to identify drug-drug interactions describing an effect. \newline  
(iii) advice: this type is used when a recommendation or advice regarding a drug-drug interaction is given. \newline 
(iv) int: this type is used when a drug-drug interaction appears in the text without providing any additional information. \newline Interactions between Leukine and other drugs have not been fully evaluated. Drugs which may potentiate the myeloproliferative effects of Leukine, such as lithium and corticosteroids, should be used with caution. & 
 
1. Leukine and lithium - effect \newline
2. Leukine and corticosteroids - effect \newline
3. Lithium and Leukine - advice \newline
4. Corticosteroids and Leukine - advice \newline 

& drug: Leukine \newline
target: lithium \newline
interaction: effect \newline

drug: Leukine \newline
target: corticosteroids \newline
interaction: effect \newline

\\ 
\midrule
HoC & Document \newline Classification & The 10 hallmark cancer taxonomy with their definitions are given below: \newline
(i) Sustaining proliferative signaling: Cancer cells can initiate and maintain continuous cell division by producing their own growth factors or by altering the sensitivity of receptors to growth factors. 
(ii) Evading growth suppressors: Cancer cells can bypass the normal cellular mechanisms that limit cell division and growth, such as the inactivation of tumor suppressor genes. 
(iii) Resisting cell death: Cancer cells develop resistance to apoptosis, the programmed cell death process, which allows them to survive and continue dividing.
(iv) Enabling replicative immortality: Cancer cells can extend their ability to divide indefinitely by maintaining the length of telomeres, the protective end caps on chromosomes. 
(v) Inducing angiogenesis: Cancer cells stimulate the growth of new blood vessels from existing vessels, providing the necessary nutrients and oxygen to support their rapid growth.
(vi) Activating invasion and metastasis: Cancer cells can invade surrounding tissues and migrate to distant sites in the body, forming secondary tumors called metastases. 
(vii) Cellular energetics: Cancer cells rewire their metabolism to support rapid cell division and growth, often relying more on glycolysis even in the presence of oxygen (a phenomenon known as the Warburg effect). 
(viii) Avoiding immune destruction: Cancer cells can avoid detection and elimination by the immune system through various mechanisms, such as downregulating cell surface markers or producing immunosuppressive signals. 
(ix) Tumor promoting inflammation: Chronic inflammation can promote the development and progression of cancer by supplying growth factors, survival signals, and other molecules that facilitate cancer cell proliferation and survival. 
(x) Genome instability and mutation: Cancer cells exhibit increased genomic instability, leading to a higher mutation rate, which in turn drives the evolution of more aggressive and drug-resistant cancer cells. \newline

Classify the following sentence in one of the above 10 hallmark cancer taxonomy. If cannot be classified, answer as "empty": \newline

 Heterogeneity in DNA damage within the cell population was observed as a function of radiation dose. & genomic instability and mutation. & genomic instability and mutation.
\\ 
\midrule
PubMedQA & Question Answering & For the question, the reference context, and the answer given below, is it possible to infer the answer for that question from the reference context? Only reply as either Yes or No or Maybe. \newline 

Question: Do some U.S. states have higher/lower injury mortality rates than others? \newline 

Reference context:  This article examines the hypothesis that the six U.S. states with the highest rates of road traffic deaths (group 1 states) also had above-average rates of other forms of injury such as falling, poisoning, drowning, fire, suffocation, homicide, and suicide, and also for the retail trade and construction industries. The converse, second hypothesis, for the six states with the lowest rates of road traffic deaths (group 2 states) is also examined. Data for these 12 states for the period 1983 to 1995 included nine categories of unintentional and four categories of intentional injury. Seventy-four percent of the group 1 states conformed to the first hypothesis, and 85\% of the group 2 states conformed to the second hypothesis. \newline   

Answer:  Group 1 states are likely to exhibit above-average rates for most other categories of injury death, whereas group 2 states are even more likely to exhibit below-average rates for most other categories of injury death. & Yes. & Yes. \\
\bottomrule
\end{tabular}}
%\vspace{-2mm}
\caption{\small{Sample prompts with ChatGPT generated responses for Relation Extraction, Document Classification, and Question Answering Tasks.}}
\label{tab:sample_prompt_response_biogpt}
\end{table*}

\begin{table*}
\centering
\tiny
\resizebox{.98\textwidth}{!}{
\setlength{\tabcolsep}{2pt} 
\begin{tabular}{p{1cm}p{1.25cm}p{8cm}p{2cm}p{2.25cm}}
\toprule
\textbf{Dataset} & \textbf{Type} &  \textbf{Prompt} & \textbf{ChatGPT Response} & \textbf{Gold Summary} \\
\midrule
iCliniq & Dialog \newline Summarization & Write a very short and concise one line summary of the following dialogue as an informal question in a healthcare forum:
\newline 
Patient: Hello doctor, I am depressed. What can I do? My parents are not supporting for a surgery which is important for my self-confidence and self-esteem. My father is not taking care of my health or well-being. Please help. Doctor: Hello. Any related incident or cause for having depression is unclear in this query. Details of surgery and symptoms which are not mentioned are appreciated to know the complete picture. Depressive feelings are generally linked with some loss, self-related issues, stress, or environmental or biological factors. It can be managed by: 1. Medicines. 2. Psychotherapy to identify the causing agents and try resolving the issue. 3. Modifying thoughts related to concerns. 4. Physical exercises like walking, jogging, and workout. 5. Diet changes - Reduce the intake of too much salt and junk food and reduce sugar intake. Consume healthy foods and fruits having serotonin and omega-3 can reduce depression.
& What are some ways to manage depression without surgery and what details should I provide to my doctor? & I am depressed, and my parents are not supporting me to overcome it. Please help
\\ 
\midrule
HealthCare Magic & Dialog \newline  Summarization & Write a very short and concise one line summary of the following dialogue as a question in a healthcare forum:
\newline
Patient: Dear Doctor,I have a 1.5 years old sick daughter, I am looking for the written file to understand some clinical numbers of a normal young child(1-2 years old):1: what is the diameter of the hepatic portal vein? 2: what is the frequency of the hepatic portal flow?I will appreciate if you could help me with these information.Kelvin Lu Doctor: 4-8 mm is range for diameter for hepatic portal vein for 1.5 year of children ,  there is no thing like frequency of hepatic portal flow.   please make it clear
  & What is the diameter of the hepatic portal vein in a 1-2 year old child, and what is the frequency of the hepatic portal flow? & What should be the diameter of hepatic portal vein of a 1.5 years old?
\\ 
\midrule
MeQSum & Question \newline  Summarization & Rewrite the following question in a short and concise form: \newline
Question: Williams' syndrome . I would like to have my daughter tested for William's syndrome. Could you please tell me where I would go or who does it in my area?  Thank you!!  & Where can I get my daughter tested for Williams' syndrome in my area? & Where can I get genetic testing for william's syndrome?
\\
\midrule
MEDIQA-QS & Question  \newline Summarization & Rewrite the following question in a short and concise form:
\newline
Question: 23 surgeries and counting......lower lip birthmark, have tried all options out the there and guess what still have it, continues to grow back.....any suggestions? Is there a cure coming in the next few years hopefully? & Any cure for a growing lower lip birthmark after 23 surgeries? & How can i get rid of a lower lip birthmark permanently?
\\
\midrule
MEDIQA-ANS & Answer \newline  Summarization & Write a very short and concise summary of the following article based on the question given below: \newline

Question: Bad Breath I have very bad breath and at times it can make myself and others sick.  I need some advice as to what I need to do. \newline

Article:  Bad Breath Halitosis Summary There are many reasons why you might have bad breath. You can get it if you don't brush and floss regularly. Bacteria that build up in your mouth and between your teeth produce the bad odor. Other problems in your mouth, such as gum disease, dry mouth or cavities, may also cause it. Sinusitis or problems with your nose may be to blame. You can also have bad breath if you eat some foods, like raw onions, garlic or cabbage. And of course smoking causes its own bad aroma. Some diseases and medicines are associated with a specific breath odor. Having good dental habits, like brushing and flossing regularly, help fight bad breath. Mouthwashes, mints or chewing gum may make your breath fresher. If you have an underlying disorder, treating it may help eliminate the breath odor. 
& The article discusses the causes and considerations of bad breath, which is usually related to poor dental hygiene. Other disorders and diseases can also produce distinct breath odors, such as fruity breath being a sign of ketoacidosis in diabetes. Proper dental hygiene, avoiding smoking, and following healthcare provider's instructions are recommended for treating bad breath. If bad breath persists or is accompanied by other symptoms, medical attention may be necessary.
  & Bad breath could be caused by poor oral hygiene, dental and  gastroesophageal diseases, infections, smoking, alcoholism, and certain foods.
\\
\midrule
MEDIQA-MAS & Answer  \newline Summarization & For the following question, some relevant answers are given below. Please write down a short concise answer by summarizing the given answers. \newline 

Question: Can I use a Power Plate machine if I have a pacemaker? \newline

Answer 1: Most machines and devices will not interfere with your pacemaker. But some with strong magnetic fields may. Always ask your provider about any specific device that you need to avoid. DO NOT put a magnet near your pacemaker.Most appliances in your home are safe to be around. This includes your refrigerator, washer, dryer, toaster, blender, computers and fax machines, hair dryer, stove, CD player, remote controls, and microwaves.You should keep several devices at least 12 inches (30 centimeters) away from the site where the pacemaker is placed under your skin. These include: Battery powered cordless tools (such as screwdrivers and drills) Plug-in power tools (such as drills and table saws) Electric lawnmowers and leaf blowers Slot machines Stereo speakers Tell all providers that you have a pacemaker before any tests are done.Some medical equipment may interfere with your pacemaker.Stay away from large motors, generators, and equipment. DO NOT lean over the open hood of a car that is running. Also stay away from: Radio transmitters and high-voltage power lines Products that use magnetic therapy, such as some mattresses, pillows, and massagers Large electrical- or gasoline-powered appliances If you have a cell phone: DO NOT put it in a pocket on the same side of your body as your pacemaker. When using your cell phone, hold it to your ear on the opposite side of your body. Be careful around metal detectors and security wands. Handheld security wands may interfere with your pacemaker. Show your wallet card and ask to be hand searched. Most security gates at airports and stores are OK. But DO NOT stand near these devices for long periods. Your pacemaker may set off alarms. After any operation, have your provider check your pacemaker. \newline

Answer 2: You will probably be able to go home after 1 day or even the same day in some cases. You should be able to return to your normal activity level quickly.Ask your provider how much you can use the arm on the side of your body where the pacemaker was placed. You may be advised not to: Lift anything heavier than 10 to 15 pounds (4.5 to 6.75 kilograms) Push, pull, and twist your arm for 2 to 3 weeks. Raise your arm above your shoulder for several weeks. When you leave the hospital, you will be given a card to keep in your wallet. This card lists the details of your pacemaker and has contact information for emergencies. You should always carry this wallet card with you. You should try to remember the name of the pacemaker manufacturer if you can in case you lose your card. \newline & It is important to ask your provider about specific devices that may interfere with a pacemaker, but most appliances in the home are safe. It is recommended to stay away from large motors, generators, and equipment, products that use magnetic therapy, and to be cautious around metal detectors and security wands. Additionally, after getting a pacemaker, some activity restrictions may apply, such as not lifting heavy objects and limiting arm movements. It is important to carry a wallet card with pacemaker details and emergency contact information. & Most machines are safe to use with a pacemaker. Some devices should be kept at least 12 inches away. These include  battery powered cordless tools (such as screwdrivers and drills) Plug-in power tools (such as drills and table saws) Electric lawnmowers and leaf blowers Slot machines Stereo speakers Some medical equipment may interfere with your pacemaker.Stay away from large motors, generators, and equipment, Radio transmitters and high-voltage power lines Products that use magnetic therapy, such as some mattresses, pillows, and massagers Large electrical- or gasoline-powered appliances. Do not put your cell phone in a pocket  or to the ear on the same side as your pacemaker.

\\
\bottomrule

\end{tabular}}
%\vspace{-2mm}
\caption{\small{Sample prompts with ChatGPT generated responses for Abstractive Summarization tasks.}}
\label{tab:sample_prompt_response_summ}
\end{table*}

\begin{table*}
\centering
\small
\begin{tabular}{lll}
\toprule
\textbf{Dataset} & \textbf{Type} & \textbf{Data Split (Train / Valid / Test)}\\
\midrule
BC5CDR & Chemical-Disease Relation Extraction & 500 / 500 / 500 \\
\midrule
KD-DTI & Drug-Target Relation Extraction & 12K / 1K / 1.3K 
\\ 
\midrule
DDI & Drug-Drug Relation Extraction & 664 / 50 / 191 
\\ 
\midrule
HoC & Document Classification & 9972 / 4947 / 4947 
\\ 
\midrule
PubMedQA & Question Answering & 450 / 50 / 500 
\\
\midrule
iCliniq & Dialog Summarization & 24851 / 3105 / 3108 
\\ 
\midrule
HealthCare Magic & Dialog Summarization & 181122 / 22641 / 22642 
\\ 
\midrule
MeQSum & Question  Summarization & 500 / - / 500  
\\ 
\midrule
MEDIQA-QS & Question  Summarization & - / 50 / 100 
\\
\midrule
MEDIQA-MAS & Answer Summarization & - / 50 / 80
\\
\midrule
MEDIQA-ANS & Answer Summarization & - / - / 552 \\
\bottomrule
\end{tabular}
\caption{{Datasets Details.}}
\label{tab:data_distribution}
\end{table*}

\end{document}